\documentclass[runningheads]{llncs}
\usepackage[T1]{fontenc}
\usepackage{algorithm}
\usepackage{algpseudocode}
\usepackage{amsmath, amssymb}
 
\usepackage{graphicx}
\usepackage{float}
\usepackage{booktabs}
\usepackage{multirow}
\usepackage{hyperref}
\usepackage{xurl}
\begin{document}
\title{SHAPCA: Consistent and Interpretable Explanations for Machine Learning Models on Spectroscopy Data}
\titlerunning{SHAPCA: Consistent Explanations for Spectroscopy Data}
\author{
Mingxing Zhang\inst{1,2,3}\orcidID{} \and
Nicola Rossberg\inst{1,2}\orcidID{0009-0005-3883-5833} \and
Simone Innocente\inst{1,4}\orcidID{0009-0002-2237-513X}\and
Katarzyna Komolibus\inst{4}\orcidID{0000-0003-4545-6100}\and
Rekha Gautam\inst{4}\orcidID{0000-0002-1176-8491}\and
Barry O'Sullivan\inst{1,2,3}\orcidID{0000-0002-0090-2085} 
Luca Longo\inst{1,3}\orcidID{0000-0002-2718-5426} \and
Andrea Visentin\inst{1,2,3}\orcidID{0000-0003-3702-4826}
}

\authorrunning{M. Zhang et al.}

\institute{
School of Computer Science and Information Technology,
University College Cork, Ireland
\and
Centre for Research Training in Artificial Intelligence,
University College Cork, Ireland
\and
Insight Centre for Data Analytics,
University College Cork, Ireland
\and
Biophotonics@Tyndall, IPIC, Tyndall National Institute, Ireland
}

\maketitle              % typeset the header of the contribution
\begin{abstract}
In recent years, machine learning models have been increasingly applied to spectroscopic datasets for chemical and biomedical analysis. For their successful adoption, particularly in clinical and safety-critical settings, professionals and researchers must be able to understand and trust the reasoning behind model predictions. However, the inherently high dimensionality and strong collinearity of spectroscopy data pose a fundamental challenge to model explainability. These properties not only complicate model training but also undermine the stability and consistency of explanations, leading to fluctuations in feature importance across repeated training runs. Feature extraction techniques have been used to reduce the input dimensionality; these new features hinder the connection between the prediction and the original signal. This study proposes \textit{SHAPCA}, an explainable machine learning pipeline that combines Principal Component Analysis (for dimensionality reduction) and Shapely Additive exPlanations (for post hoc explanation) to provide explanations in the original input space, which a practitioner can interpret and link back to the biological components.  The proposed framework enables analysis from both global and local perspectives, revealing the spectral bands that drive overall model behaviour as well as the instance-specific features that influence individual predictions. Numerical analysis demonstrated the interpretability of the results and greater consistency across different runs.

\end{abstract}
\section{Introduction}
\label{sec:introduction}
Spectroscopy techniques are broadly employed in the scientific and industrial domains, including biomedical research, chemistry, materials science, and environmental monitoring. Techniques such as Raman, infrared, near-infrared, and nuclear magnetic resonance spectroscopy provide rich molecular fingerprint information, enabling non-destructive and label-free analysis of samples \cite{esmonde2022role}. In recent years, the increased availability of high-resolution spectroscopy instruments has enabled the generation of large, complex datasets \cite{parker2025analysis}. Machine learning techniques have been increasingly used to analyse and evaluate spectroscopic data for classification, regression, and anomaly detection \cite{rossberg2025machine}. In particular, supervised learning models have demonstrated outstanding performance in spectroscopic classification, effectively discriminating between different chemical compositions, material states, and disease conditions based on spectral features \cite{liu2026recent}. Looking forward, the deep integration of spectroscopy and machine learning is expected to play an increasingly important role in automated decision-making systems, diagnostics, and large-scale analytical pipelines \cite{bertazioli2024integrated}.

Despite the strong predictive performance of machine learning models applied to spectroscopic data, their adoption in high-stakes domains remains limited, particularly in medical and clinical applications. In this high-risk field, the European Union Artificial Intelligence Act (EU AI Act) now imposes legally binding requirements for transparency, accountability, and meaningful human oversight in AI-driven decision-making systems \cite{union2024regulation}. At the same time, clinicians and healthcare practitioners must be able to understand and trust the reasoning behind algorithmic predictions in order to safely integrate such systems into clinical practice \cite{yang2022explainable}. As a result, eXplainable Artificial Intelligence (XAI) has become a fundamental requirement for the responsible deployment of machine learning models, enabling predictions to be interpreted, validated, and justified in line with ethical and regulatory expectations.

However, achieving explainability in spectroscopic machine learning models is particularly challenging due to the intrinsic characteristics of spectroscopic data. Spectra are typically high-dimensional, comprising hundreds of features, and exhibit strong collinearity, as adjacent wavelengths originate from the same molecular vibrations and spectral peaks are inherently broad and overlapping \cite{guo2025artificialintelligencespectroscopyadvancing}. As a result, information relevant to classification is often distributed across extended spectral regions rather than confined to isolated peaks \cite{pandiselvam2022recent}. These properties complicate both model training and post-hoc explanation: feature importance measures may become diluted across correlated variables and exhibit instability or inconsistency across repeated training runs \cite{salih2024explainableartificialintelligencedependent}. Consequently, standard explainability techniques, when directly applied to raw spectroscopic features, may fail to yield reliable or physically meaningful interpretations. Feature extraction approaches, such as Principal Component Analysis (PCA), are widely used in spectroscopy to project the spectra, reducing correlation and noise. However, their components do not have a direct biological meaning, and they can limit the applicability of XAI approaches since they would explain the decisions in relation to the values in the reduced space.

To address the issues presented above, this paper introduces \textit{SHAPCA}, a model-agnostic post-hoc explanation algorithm that captures spectral correlation structure by grouping highly correlated wavelengths into a low-dimensional latent representation used by a classifier and reprojects their impact into the original input space, improving practitioners' understanding. Model training and classification are conducted in the latent space, reducing computational complexity while maintaining (or improving) predictive accuracy. We then compute Shapley Additive exPlanations (SHAP) values \cite{lundberg2017unifiedapproachinterpretingmodel} at the component level to quantify the contribution of each latent component to the prediction. By performing explanation in the latent space, contributions are less susceptible to dilution across individual correlated features and exhibit improved stability across repeated runs. Finally, component-level contributions are back-projected to the original wavenumber axis to yield feature-space explanations that are directly interpretable and suitable for practical deployment. The explanation is superimposed over the original signal, highlighting which areas are relevant for the machine learning decision-making and if they are higher or lower than the other samples. 

The structure of the paper is as follows: Section \ref{sec:literature} positions this work in the relevant literature by providing an overview of the existing XAI approaches for spectroscopy data and of combinations of PCA and SHAP values for explainability. The datasets used, and the proposed approach are presented in Section \ref{sec:methodology}. Section \ref{sec:results} presents and discusses the numerical analysis. Finally, Section \ref{sec:conclusion} summarises the contributions and findings. 

\section{Literature Review} \label{sec:literature}
To contextualize this work, previous XAI applications to spectroscopic data are reviewed and evaluated. Additionally, the integration of PCA and SHAP in the domain of XAI is analysed.

\subsection{XAI for spectroscopy}

As the field of machine learning (ML) continues to grow, it is being increasingly integrated into the field of spectroscopy, permitting the automation of a range of predictive tasks. The applications of ML to spectroscopy have been especially impactful in fields such as chemistry \cite{westermayr2025machine,anker2023machine,tetef2021unsupervised}, agriculture \cite{su2020advanced,farber2022raman,mokere2025soil,peng2023estimation}, and biomedicine \cite{zhao2023potential,rossberg2025machine,nogueira2024diffuse,rodriguez2023evaluating}. These results demonstrate that machine learning models are capable of successfully classifying high-dimensional spectroscopy signals, demonstrating strong potential for the automation of a range of important tasks. However, one major concern with the implementation of ML models, especially in critical domains such as medicine, is their black-box nature and the associated inability to reliably and faithfully explain decision processes. In addition to being a policy requirement under the European Union AI act, the implementation of explainability is also vital to prevent problems such as model bias and shortcut learning and promote practitioner trust \cite{thalpage2023unlocking,kastner2021relation,mensah2023artificial}. 

A wide range of XAI techniques have been proposed in the domain of ML, accounting for various data, model, and explanation types. However, in conjunction with spectroscopic signals, the implementation of XAI is challenging due to the high-dimensional and intercorrelated nature of these signals. As a result, traditional explanatory methods frequently yield explanations that are inconsistent across model iterations and consequentially unreliable. A comprehensive review of existing explanatory approaches for spectroscopic signals is provided by Contreras et al.~\cite{contreras2025explainable}. Their review indicates that relatively few XAI studies have been specifically developed for spectroscopy. Among the existing work, explanation methods can generally be categorized into four groups: activation-based, perturbation-based, gradient-based, and surrogate-based approaches, as well as various combinations of these techniques.

Rossberg et al.~\cite{rossberg2025explainable} proposed two model-specific XAI-based feature-selection methods for Raman spectroscopy, combining CNNs with Grad-CAM and Transformers with attention-derived importance scores. Their results showed that the CNN-Grad-CAM approach achieved the highest average accuracy, although no single method consistently outperformed all others across every scenario. Despite this strong predictive performance, both approaches are tied to specific model architectures, and the selected features remain mainly individual wavenumbers, which may limit physical interpretability in spectroscopy, where variables are highly correlated and meaningful information is often distributed across broader spectral regions rather than isolated features.

While Contreras et al.~\cite{contreras2024spectral} proposed a spectral-zones-based SHAP and LIME framework for deep learning on spectral data, in which neighboring wavelengths are grouped into contiguous zones before perturbation so that the explanations emphasize interpretable spectral regions rather than single features. However, this strategy remains constrained by spatial adjacency: it can only group neighboring wavelengths and therefore cannot capture correlated but non-adjacent features that may originate from the same chemical component. As noted in the study, some biochemical signatures, such as fat-related signals, may appear in several separate spectral regions, meaning that chemically related but spatially distant variables are still treated independently.

In addition, \cite{wang2021machine} developed an ANN-based model to predict solution conductivity from plasma optical emission spectra and they extract the most dominent emission lines from each spectrum and employed LIME to identify influential ones for the model’s predictions. Their results showed that OH, H$_\gamma$ and H$_\beta$ emission lines were critical to the model’s decisions, and that these differed from the lines typically selected by human experts. This further highlights the value of grouping spectral features for explainability, but also suggests that more principled grouping strategies are needed.

Overall, these studies demonstrate that some form of feature grouping or segmentation is essential for explaining high-dimensional spectroscopy data. However, most of the existing approaches still rely mainly on architectural constraints or local spectral adjacency, and therefore may fail to capture the broader correlation structure of the data. In our study, we address this limitation by applying PCA to group correlated features into latent components, providing a more principled way to represent jointly varying spectral information and enabling explanations at a level that is more robust and potentially more meaningful than isolated feature-wise attribution.

\subsection{PCA for high-dimensional data}

One approach to combat the high dimensionality and inter-correlation of spectroscopic data is the implementation of PCA. PCA generates new variables based on the directions of maximum variance within a dataset. As the generated components are perpendicular to each other within the feature space, they are uncorrelated and hence can be more reliably classified and explained through machine learning systems. In addition, the implementation of PCA for spectroscopic signals was found to reveal systematic variation and underlying patterns in complex spectroscopic datasets \cite{wiklund2007spectroscopic}. However, one drawback of the implementation of PCA is the associated loss of information linking the ML output to the original input variables. This, in turn, decreases transparency and hinders model implementation, as practitioners are unable to link explanations to spectral signatures. As such, the ability to decompose components back to the spectral inputs after ML classification would permit the successful integration of XAI techniques for spectroscopic data. 

Some previous research has investigated the implementation of PCA in combination with XAI techniques in other fields that also have high-dimensional datasets, since there is not much similar work for spectroscopy data. Chen et al.~\cite{chen2025sparse} proposed an interpretable stacked ensemble framework for predicting SO\(_2\) concentration in coal-fired boilers. After applying the Relief regression algorithm to select informative variables, they used Sparse PCA to reduce redundancy in the high-dimensional input features and constructed a base-model pool consisting of 17 machine learning models, which were then combined through stacking strategies. SHAP was subsequently employed to analyse the contribution of input variables to the model predictions. Although Sparse PCA and SHAP were both used within the framework, their integration was relatively loose, as Sparse PCA mainly served as a preprocessing step for dimensionality reduction rather than being directly coupled with the explanation stage. Nevertheless, the study still illustrates the respective advantages of Sparse PCA for handling high-dimensional data and SHAP for improving model interpretability. Other fields have successfully implemented PCA in conjunction with SHAP for applications such as the diagnosis of prostate cancer \cite{b2024innovative} and brain tumours \cite{happila2025enhancing}.

In this study, for spectroscopy data, we extend these efforts by integrating Sparse PCA with SHAP and explicitly projecting all attribution scores back to the original wavenumber space. This design preserves the interpretability advantages of sparsity while enabling feature-level explanations that remain linked to physically meaningful spectral bands, allowing the importance of original spectral variables to be quantified in a stable and scientifically interpretable manner.

\section{Methodology}
\label{sec:methodology}

This section describes the dataset used in this study, the proposed analysis pipeline, the back-projection steps applied, and the consistency-checking procedures used to evaluate the results. Together, these parts establish the study's logical and methodological foundation.

\subsection{Data Preparation}
We selected two datasets to evaluate the approach on both binary and multi-class classification tasks. To examine the behaviour of SHAP explanations in both settings, two datasets, collected using different spectroscopic modalities, are considered.

\noindent \paragraph{Raman Spectroscopy Dataset.} This dataset was collected by \cite{maryam2025liquid} and consists of Raman spectroscopy measurements acquired from cheek mucosa tissue for cancer diagnosis. The data consists of 484 Raman spectra obtained from 77 patients, with each spectrum assigned a class label corresponding to the underlying tissue condition, resulting in three diagnostic classes: Potentially Malignant Healthy(PMH), Potentially Malignant Lesion (PML), and Healthy(H). Raman intensity values were recorded over the spectral range of 398.75–1998.75 cm{$^{-1}$}, forming high-dimensional feature vectors that capture the biochemical characteristics of the cheek mucosa tissue.

Prior to classification, the following preprocessing steps were taken:
\begin{itemize}
  \item Samples labelled `PMH' were excluded. Since this study focuses on distinguishing healthy from malignant cases, and PMH samples do not clearly fit into either category, this class was removed to maintain a clear binary classification.
  \item Cropped the spectral region below 411.25 cm$^{-1}$, as this range was dominated by noise.
  \item Applied Savitzky--Golay smoothing with a window length of 5 and a polynomial order of 2 along the spectral dimension.
  \item Performed baseline correction using the DRPLS algorithm with parameters $\lambda = 5 \times 10^{5}$ and $p = 0.003$.
  \item Applied maximum intensity normalization on each spectrum.
  \item The training and testing sets were split at the patient level to prevent data leakage.
\end{itemize}

\noindent \paragraph{DRS Dataset.} This dataset was collected by \cite{li2024extended} and comprises 5,215 spectra spanning six tissue classes commonly encountered in orthopaedic surgeries: Bone Cement (boneCement), Bone Marrow (boneMarrow), Cartilage (cartilage), Cortical Bone (cortBone), Muscle (muscle), and Trabecular Bone (traBone). Spectra were acquired over the range  $355.016$--$1849.739~\mathrm{nm}$. Preprocessing was performed in accordance with the procedure described in the original study, prior to applying the proposed pipeline.

\subsection{Pipeline}

This study proposes an interpretable machine-learning pipeline for spectroscopic data that integrates PCA, supervised classification, and SHAP-based explainability. The procedure comprises the following steps:
\begin{enumerate}
    \item PCA decomposition: the input spectrum is transformed into a low dimensional representation by projecting it through the PCA loadings $W$, yielding principal component (PC) values;
    \item Supervised prediction: the PC values are used as inputs to a classifier to produce class predictions; in this work, we used Random Forest(RF) and Support Vector Classifiers (SVC);
    \item  SHAP explainability: Shapley values $\Phi$ are computed for the latent components to quantify their contributions to the model output;
    \item Back-projection and visualisation: the Shapley values' contribution is reprojected into the original spectra as opacity to highlight the parts of the input that are more relevant to the decision. Their being higher or lower than the rest of the samples is driven by the PC values, which give the color red for higher and blue for lower. Combining and overlaying these two aspects onto the original spectral shape to yield physically interpretable, feature-level explanations.
\end{enumerate}

\begin{figure}[H]
    \centering
    \includegraphics[width=\textwidth]{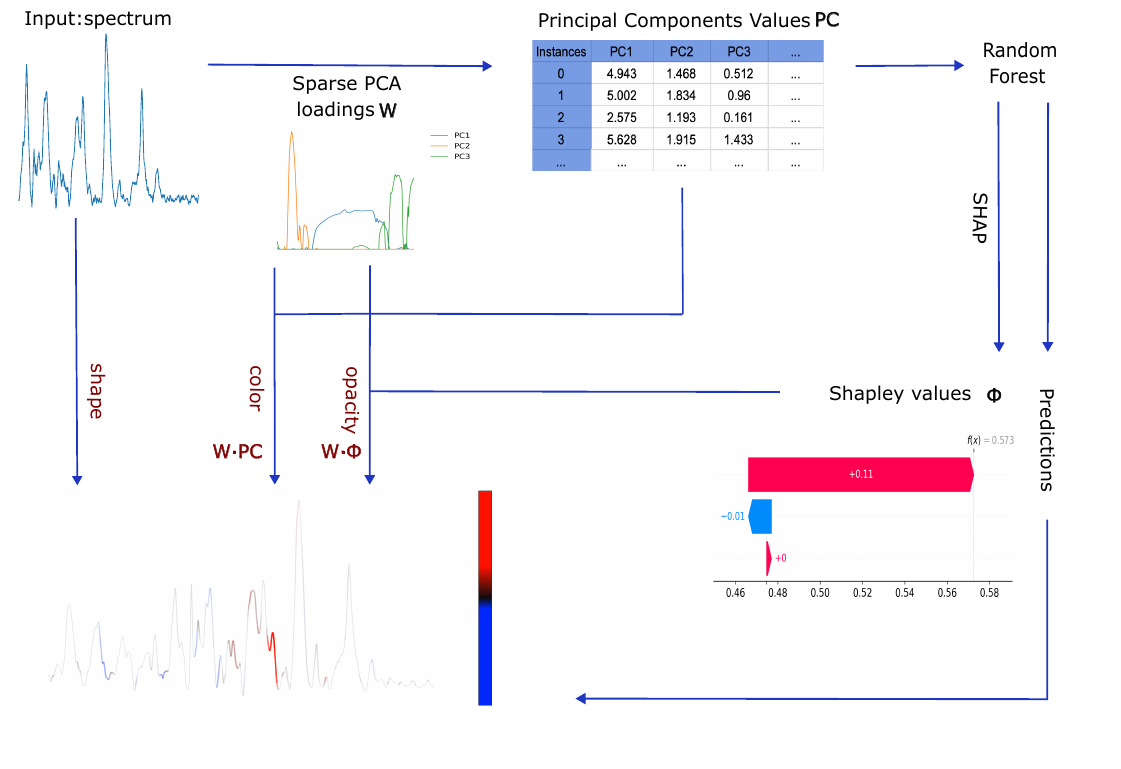}
    \caption{
    Step-by-step workflow of the proposed interpretable spectroscopic pipeline.    \label{fig:pipeline}}

\end{figure}

Figure ~\ref{fig:pipeline} shows a summary of the pipeline. The individual components and steps are described in the following Sections.

\subsubsection{Sparse PCA}

As discussed in Section \ref{sec:introduction}, individual molecular vibrations influence a range of neighbouring wavenumbers, and multiple biochemical components contribute overlapping signals within the same spectral regions, resulting in strong collinearity among spectroscopic features. PCA is well-suited to capture this covariance structure because it identifies orthogonal directions in the data that explain the greatest variance. In spectroscopic data, groups of correlated variables often vary together and therefore contribute jointly to these dominant directions of variation. The resulting principal components offer an approximate representation of the main patterns present in the spectra, although they are derived by maximizing variance rather than explicitly modelling collinearity itself. Nevertheless, fully disentangling the latent structure underlying spectroscopic data is a highly challenging problem and is beyond the scope of the present study.

Given a dataset consisting of $p$ original features (in this case, wavenumbers), PCA constructs a new set of orthogonal variables, referred to as principal components, which are ordered according to the amount of variance they explain. Each principal component is defined as a linear combination of the original features. Specifically, the $k$-th principal component can be written as
\begin{equation}
PC_k = \sum_{j=1}^{p} w_{jk} x_j,
\end{equation}
where $x_j$ denotes the $j$-th original feature and $w_{jk}$ represents the corresponding loading. The vector of loadings
\[
\mathbf{w}_k = (w_{1k}, w_{2k}, \ldots, w_{pk})^\top
\]
defines the direction of the $k$-th principal component in the original feature space.

The loading vectors quantify the contribution of each original feature to a given component. Features with larger absolute loading values exert a stronger influence on the component, whereas features with near-zero loadings contribute minimally. As such, the loadings provide a direct link between the transformed representation and the original variables, enabling interpretation of principal components in terms of physically meaningful structures, such as spectral bands associated with underlying biochemical variations.

However, in standard PCA, each principal component is typically expressed as a linear combination of all original wavenumbers, even if many of the corresponding loadings have very small magnitudes. This results in dense components that are difficult to interpret, particularly for spectroscopic data, where diagnostically relevant information is often localised to specific biochemical bands rather than distributed across the entire spectrum.

To concentrate the explanations on the spectral regions most relevant to the model’s decision-making process, we therefore employed Sparse Principal Component Analysis (Sparse PCA). Sparse PCA extends conventional PCA by imposing sparsity constraints on the component loadings, forcing many coefficients to be exactly zero while retaining non-zero contributions for only a limited subset of wavenumbers. This sparsity property allows each component to be associated with a small number of informative spectral regions, rather than diffuse contributions across the full spectral range. Consequently, Sparse PCA yields more interpretable component representations in terms of localised spectral features or biochemical bands, while still preserving the ability to capture the dominant patterns of variation in the data.

\subsubsection{Classification}
Sparse PCA component values are used as inputs to the downstream classifier. In our experiments, Random Forest and Support Vector Classifiers (SVC) are chosen.

Importantly, RF is particularly well-suited for SHAP-based interpretation. SHAP provides an exact and computationally efficient TreeExplainer for tree-based models, enabling stable and theoretically grounded Shapley value estimation. In contrast, SHAP explanations for non-tree-based classifiers, such as SVC, typically rely on model-agnostic explainers; we applied Kernel Explainer, which are computationally expensive and less stable in high-dimensional settings. Moreover, SVCs do not natively produce calibrated probability estimates, requiring additional probability calibration, which introduces further approximation and uncertainty. By contrast, Random Forests provide native probabilistic outputs through ensemble averaging, allowing SHAP values to be computed directly and consistently.

\subsubsection{Train/test configuration and hyperparameters selection}

The Raman dataset was first split at the patient level into an 80/20 train–test partition to prevent information leakage across patients, whereas the DRS dataset was split at the sample level. Hyperparameter tuning of the pipeline was performed exclusively on the training portion using a two-stage strategy: an initial randomized search with cross-validation to identify stable regions of the hyperparameter space, followed by a grid search for fine-grained refinement. GroupKFold was used for the Raman dataset and stratified cross-validation was used for the DRS dataset. After tuning, the final model was retrained on the full training set and evaluated once on the held-out test split.

Because overly restrictive sparsity constraints in Sparse PCA can reduce classification performance, hyperparameters of both Sparse PCA and the downstream classifier were jointly optimized. Candidate models were evaluated based on predictive accuracy and the sparsity of the PCA components. Among configurations with comparable accuracy, preference was given to models exhibiting higher sparsity to enhance interpretability, thereby ensuring a balanced trade-off between classification performance and clarity of feature-level explanations. For both the Raman and DRS datasets, the selected Sparse PCA configuration retained 10 components, with the sparsity parameter $\alpha = 1$ for both the RF and SVC based pipelines.

\subsubsection{SHAP}

After classification, all model predictions were provided to SHAP for explanation. SHAP is a post-hoc explainability method grounded in cooperative game theory. It quantifies the contribution of each feature (in this study, each latent component) to a model’s prediction by computing its average marginal contribution across all possible feature subsets \cite{lundberg2017unifiedapproachinterpretingmodel}. Under the additive explanation framework, the prediction for an instance 
$x$ is decomposed as
\begin{equation}
f(x) = \phi_0 + \sum_{k=1}^{K} \phi_k(PC_k),
\end{equation}
where $K$ is the number of components, $\phi_k(PC_k)$ denotes the contribution of component $k$, and $\phi_0$ is a baseline term, commonly defined as the expected model output over a reference (background) distribution:
\begin{equation}
\phi_0 = \mathbb{E}[f(X)].
\end{equation}

Shapley values represent additive, sample-wise contributions defined with respect to a common baseline, which makes them comparable across samples and amenable to meaningful aggregation. This property distinguishes SHAP from other attribution methods, such as gradient-based saliency maps or attention weights, whose values are often sensitive to input scaling, lack a consistent reference across samples, and therefore do not support semantically reliable aggregation.\cite{lundberg2020local}.

Importantly, for the DRS dataset, which involves a multi-class classification task, component attributions are inherently class-conditional, as the model learns distinct decision functions for each output class. Accordingly, SHAP computes component contributions separately for each class, yielding class-specific Shapley values that quantify how each component influences different class predictions. As a result, global explanations must be derived in a class-wise manner to preserve semantic meaning. Aggregating attributions across classes would mix contributions that correspond to different class outputs, thereby blurring or even cancelling class-specific evidence and obscuring the distinct decision logic learned for each class. By contrast, aggregating SHAP values within each class preserves the class-conditional semantics of the attributions and yields global patterns that remain interpretable and directly aligned with the multi-class classification objective.

\subsection{Visualization}
The goal of the visualisation is to enable practitioners to interpret the results in terms of biologically meaningful spectral regions. In particular, the visualisation aims to answer two key questions of practical relevance: which regions of the spectrum contribute to the model’s decision, and whether these regions exhibit higher or lower intensity than expected.

\begin{figure}[!t]
    \centering
    \includegraphics[width=\textwidth]{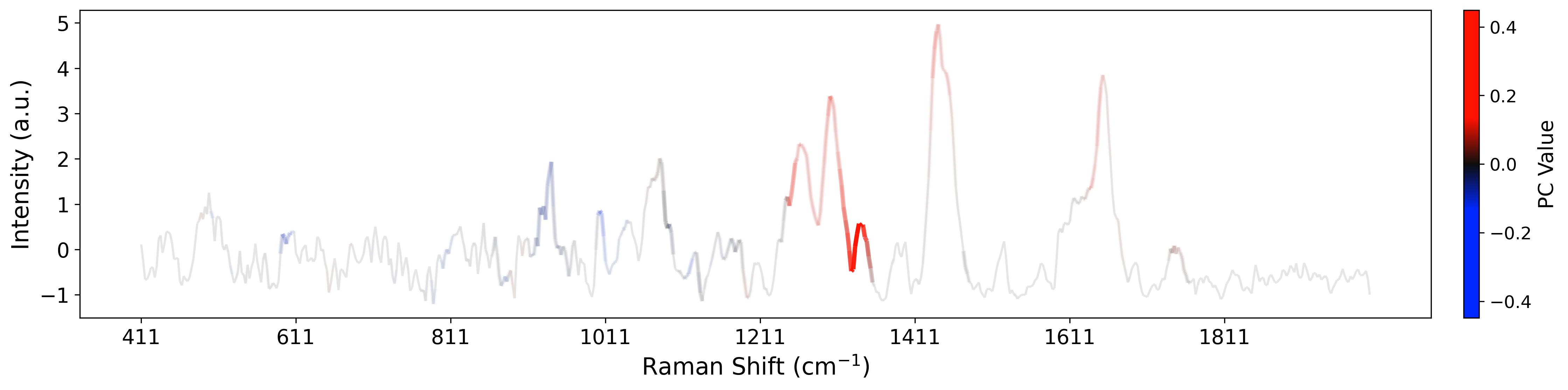}
    \caption{
      A Raman spectroscopy sample. Opacity indicates the most decisive spectral regions for the prediction, while colour encodes intensity relative to the dataset average (red = higher, blue = lower).
    }
    \label{fig:example_explanation}
\end{figure}

To this end, two component-level quantities are back-projected to the original feature space: (i) component-wise Shapley values, which are mapped to opacity (alpha) to indicate the strength of contribution to the prediction, and (ii) component values, which are encoded by colour to represent their relative magnitude across classes. This joint encoding enables simultaneous assessment of where the model draws evidence from and whether the corresponding spectral features are higher or lower than expected. Figure \ref{fig:example_explanation} shows an example of a spectra explanation. In this example, the peak around 1350 Raman shift is the most impactful for the classification, and it is higher than normal. 

\subsubsection{Global Explanation} 
Global explanation aims to characterize the overall decision behaviour of the trained model by identifying spectral regions that consistently contribute to class discrimination across the dataset. Rather than focusing on individual samples, this analysis aggregates model explanations at the class level, thereby reducing the influence of sample-specific noise and highlighting stable, representative patterns.
As shown in Algorithm~\ref{alg:global explanation}, for each class, SHAP values are extracted and averaged across samples predicted as that class to obtain a mean component-level contribution vector $\bar{\boldsymbol{\phi}}^{(c)}$. These component-wise importance scores are then back-projected to the original wavenumber space using the absolute Sparse PCA loadings $\mathbf{W}$, yielding a class-specific feature importance vector $\boldsymbol{\Psi}^{(c)}$:

\begin{equation}
\boldsymbol{\Psi}^{(c)}
\;=\;
\big(\bar{\boldsymbol{\phi}}^{(c)}\big)^{\top} \, {W}.
\end{equation}

In Figure \ref{fig:pipeline}, this encodes the opacity (transparency) of the explanation superimposed on the original spectra.

\begin{algorithm}[!t]
\caption{Class-wise SHAP Aggregation and Back-Projection to Feature Space}
\label{alg:global explanation}
\begin{algorithmic}[1]

\Statex {Input:}
\Statex \hspace{1.5em} $\boldsymbol{\Phi} \in \mathbb{R}^{N \times K \times C}$: SHAP values
\Statex \hspace{3em} ($N$: number of samples, $K$: number of components, $C$: number of classes)
\Statex \hspace{1.5em} $\hat{\mathbf{y}} \in \{1,\dots,C\}^N$: predicted class label for each sample
\Statex \hspace{1.5em} $\mathbf{W} \in \mathbb{R}^{K \times P}$: Sparse PCA loadings
\Statex \hspace{3em} ($P$: number of original features / wavenumbers)

\Statex {Output:}
\Statex \hspace{1.5em} $\boldsymbol{\Psi}^{(c)} \in \mathbb{R}^{P}$: feature-level importance vector for each class $c$

\For{each class $c \in \{1,\dots,C\}$}
    \State Extract SHAP values for class $c$:
    \Statex \hspace{1.5em} $\boldsymbol{\Phi}^{(c)} \in \mathbb{R}^{N \times K}$

    \State Select samples predicted as class $c$

    \State Compute the mean SHAP value of each component over these samples:
    \Statex \hspace{1.5em} $\bar{\boldsymbol{\phi}}^{(c)} \in \mathbb{R}^{K}$

    \State Initialise feature-level importance vector:
    \Statex \hspace{1.5em} $\boldsymbol{\Psi}^{(c)} \gets \mathbf{0} \in \mathbb{R}^{P}$

    \For{each component $k$ in $\{1,\dots,K\}$}
        \State Distribute component importance to original features:
        \Statex \hspace{1.5em}
        $\boldsymbol{\Psi}^{(c)} \gets
        \boldsymbol{\Psi}^{(c)} + \bar{\phi}^{(c)}_k \cdot|\mathbf{W}_{k}|$
    \EndFor
\EndFor

\State \textbf{return} $\{\boldsymbol{\Psi}^{(c)}\}_{c=1}^{C}$

\end{algorithmic}
\end{algorithm}

As for component values, they are first normalized across samples and then averaged within each predicted class to obtain class-wise normalized component profiles $\bar{\boldsymbol{pc}}^{(c)}$. These profiles are subsequently projected back to the original feature space using the Sparse PCA loadings $\mathbf{W}$, and the contributions from all normalised components are summed to yield a class-specific projection vector $\boldsymbol{PC}^{(c)}$:

\begin{equation}
\boldsymbol{PC}^{(c)}
= \sum_{k=1}^{K} \big(\bar{\boldsymbol{pc}}^{(c)}\big)_k \,\mathbf{W}_k
\label{eq:class_component_backproj}
\end{equation}

This highlights whether the specific part of the spectra encoded by the component is higher or lower compared to the other datapoints. The information is encoded through color (Figure \ref{fig:pipeline}) with a gradient similar to SHAP beeswarm plots.

\subsubsection{Local Explanation}

To explain a given spectrum for a target class, the component-level SHAP values were separated into positive and negative parts to preserve their signs. The approach is summarised in Algorithm \ref{alg:local_shap}. Positive contributions correspond to components that support the class prediction, whereas negative contributions indicate suppressive effects.  These two parts were then independently back-projected to the original feature space using the absolute Sparse PCA loadings $\mathbf{W}$, producing separate spectral contribution maps, $\boldsymbol{\Psi}_i^{+}$ and $\boldsymbol{\Psi}_i^{-}$, that highlight class-supporting and class-opposing evidence, respectively.

\begin{equation}
\boldsymbol{\Psi}_i^{+}=\big(\boldsymbol{\phi}_i^{+}\big)^{\top}\,\mathbf{W},
\qquad
\boldsymbol{\Psi}_i^{-}=\big(\boldsymbol{\phi}_i^{-}\big)^{\top}\,\mathbf{W}.
\end{equation}

\noindent where $i$ denotes the $i$-th instance. $\boldsymbol{\Psi}_i^{+}$ represents the features that are supporting the class prediction of instance $i$, while $\boldsymbol{\Psi}_i^{-}$ the features of the negative contribution.

\begin{algorithm}[!t]
\caption{Sample-wise SHAP and Back-projection to Feature Space}
\label{alg:local_shap}
\begin{algorithmic}[1]

\Statex {Input:}
\Statex \hspace{1.5em} SHAP values $\boldsymbol{\Phi}$
\Statex \hspace{1.5em} spectrum index $i$
\Statex \hspace{1.5em} spectrum predicted class $c$
\Statex \hspace{1.5em} Sparse PCA loadings $\mathbf{W}$

\Statex {Output:}
\Statex \hspace{1.5em} Feature-level contribution maps $\boldsymbol{\Psi}^{+}, \boldsymbol{\Psi}^{-}$

\State Extract component-level SHAP values for spectrum $i$ and class $c$:
\Statex \hspace{1.5em} $\boldsymbol{\phi}^{(c)}_i \in \mathbb{R}^{K}$

\State Separate positive and negative SHAP contributions:
\Statex \hspace{1.5em}
$\phi_{i,k}^{+} =
\begin{cases}
\phi_{i,k}^{(c)}, & \phi_{i,k}^{(c)} > 0 \\
0, & \text{otherwise}
\end{cases}$

\Statex \hspace{1.5em}
$\phi_{i,k}^{-} =
\begin{cases}
\phi_{i,k}^{(c)}, & \phi_{i,k}^{(c)} < 0 \\
0, & \text{otherwise}
\end{cases}$

\State Initialise feature-level intensity vectors:
\Statex \hspace{1.5em} $\boldsymbol{\Psi}^{+}, \boldsymbol{\Psi}^{-} \gets \mathbf{0}$

\For{each component $k$ in $\{1,\dots,K\}$}
    \State $\boldsymbol{\Psi}^{+} \gets \boldsymbol{\Psi}^{+} + \phi^{+}_{i,k} \cdot |\mathbf{W}_{k}|$
    \State $\boldsymbol{\Psi}^{-} \gets \boldsymbol{\Psi}^{-} + \phi^{-}_{i,k} \cdot |\mathbf{W}_{k}|$
\EndFor

\State {return} feature-level positive and negative contribution maps $\boldsymbol{\Psi}^{+}$ and $\boldsymbol{\Psi}^{-}$

\end{algorithmic}
\end{algorithm}

For the component values, a similar back-projection procedure to that used in the global explanation was applied. Specifically, the normalised component values of the target instance $\boldsymbol{pc}_k$ were multiplied by the corresponding Sparse PCA loadings $\mathbf{W}_k$ and summed, thereby projecting the instance-level component representation back to the original feature space:

\begin{equation}
\mathbf{PC}
= \sum_{k=1}^{K} \boldsymbol{pc}_k\,\mathbf{W}_k
\label{eq:local_component_backproj}
\end{equation}

where $\boldsymbol{PC}$ encodes the color that relate spectrum $i$ to the other spectra. The color component is used both for the positive and negative explanations.

\subsection{Consistency Check}

To assess the stability and consistency of the explanation methods, both the baseline pipeline (SHAP) and the proposed pipeline (SHAPCA) were trained under a five-fold cross-validation protocol, yielding five independently trained models per method. For each trained model, global and local explanations were obtained by back-projecting the corresponding Shapley values to the original spectral domain and evaluating these projections on an identical held-out test set. Explanation stability was measured using pairwise similarity and correlation between the five SHAP models and between the five SHAPCA models, for both global and local projections. The resulting consistency scores were then averaged within each method and compared between SHAP and SHAPCA to determine whether the proposed pipeline produces more reproducible explanations across folds. This approach has been inspired by \cite{gwinner2024comparing}.

Cosine similarity and Pearson correlation are adopted in this study. Cosine similarity quantifies the directional alignment between two vectors and is therefore primarily sensitive to the overall pattern (i.e., the relative distribution of peaks and valleys) of the back-projected Shapley vectors rather than their absolute magnitudes. For two back-projected Shapley vectors $\boldsymbol{\Psi}_m$ and $\boldsymbol{\Psi}_n$, cosine similarity is defined as
\begin{equation}
\mathrm{cos\_sim}(\boldsymbol{\Psi}_m,\boldsymbol{\Psi}_n)
=
\frac{\boldsymbol{\Psi}_m^{\top}\boldsymbol{\Psi}_n}{\lVert \boldsymbol{\Psi}_m\rVert_2 \, \lVert \boldsymbol{\Psi}_n\rVert_2},
\end{equation}
where $\boldsymbol{\Psi}_m^{\top}\boldsymbol{\Psi}_n$ denotes the dot product and $\lVert\cdot\rVert_2$ denotes the Euclidean norm. By construction, $\mathrm{cos\_sim}(\boldsymbol{\Psi}_m,\boldsymbol{\Psi}_n)\in[-1,1]$, with values close to $1$ indicating highly aligned explanation patterns, values near $0$ indicating weak alignment, and values close to $-1$ indicating opposing patterns.

Pearson correlation complements cosine similarity by measuring how strongly two explanation vectors co-vary after removing their mean offsets. In practice, this indicates whether two back-projected attribution maps exhibit peaks and valleys at consistent spectral locations, independent of any constant baseline shift. This is particularly relevant for spectroscopy explanations because back-projected Shapley vectors obtained from different folds may differ by an additive offset (or overall centering) even when the underlying discriminative pattern is preserved. For two back-projected Shapley vectors $\boldsymbol{\Psi}_m$ and $\boldsymbol{\Psi}_n$ with $p$ spectral features, Pearson correlation is defined as
\begin{equation}
\rho(\boldsymbol{\Psi}_m,\boldsymbol{\Psi}_n)
=
\frac{\sum_{i=1}^{p}\left(\Psi_{m,i}-\bar{\Psi}_m\right)\left(\Psi_{n,i}-\bar{\Psi}_n\right)}
{\sqrt{\sum_{i=1}^{p}\left(\Psi_{m,i}-\bar{\Psi}_m\right)^2}\;
 \sqrt{\sum_{i=1}^{p}\left(\Psi_{n,i}-\bar{\Psi}_n\right)^2}},
\end{equation}
where $\Psi_{m,i}$ denotes the $i$-th element of $\boldsymbol{\Psi}_m$, and $\bar{\Psi}_m=\frac{1}{p}\sum_{i=1}^{p}\Psi_{m,i}$ (analogously, $\bar{\Psi}_n$) is the mean of the corresponding explanation vector. The coefficient satisfies $\rho\in[-1,1]$, where values close to $1$ indicate strong agreement in the mean-centred spectral pattern, values near $0$ indicate weak linear association, and negative values indicate opposing patterns. Reporting both cosine similarity and Pearson correlation, therefore, provides a more complete assessment of explanation stability.

Distance-based metrics such as Euclidean distance have been used in related work~\cite{KaragozOzcelebiMeratnia2025BenchmarkingXAI}; however, they are not well suited to the present setting, since the SHAP provides $\boldsymbol{\Psi}$ values for a limited number of features, their average intensity is low and many values are 0, leading to smaller errors overall.

\section{Results and Discussion}
\label{sec:results}
This section presents the pipeline performance across the two datasets. In each application, we computed the global and local explanations and investigated the biological components associated with them. Then, the consistency of the explanation across different iterations for SHAPCA and the baseline SHAP is analysed. For the sake of reproducibility, the code used in this study is available online \footnote{\href{https://github.com/appleeye007/SHAPCA}{https://github.com/appleeye007/SHAPCA}}.

Table~\ref{tab:average_performance_raman_drs} summarises the average predictive performance on both datasets. Each pipeline was evaluated over 100 independent runs, and the results are reported in terms of mean test accuracy and F1 score, together with their 95\% confidence intervals. Overall, across both datasets, introducing Sparse PCA led to a decrease in both accuracy and F1 score compared with using the original feature space. Although the two classifiers performed at a broadly similar level on both datasets, RF was selected as the reference model for the global and local explanation visualisations because it enables the use of TreeExplainer and yielded more stable explanations across repeated runs, as shown later in the consistency analysis.

\begin{table*}[!t]
\centering
\caption{Mean test accuracy and F1 score for Raman spectroscopy and DRS data, reported as mean $\pm$ 95\% confidence interval.}
\label{tab:average_performance_raman_drs}

\setlength{\tabcolsep}{5pt}  
\renewcommand{\arraystretch}{1.15}

\begin{tabular}{lcccc}
\hline
\multirow{2}{*}{\textbf{Pipeline}} 
& \multicolumn{2}{c}{\textbf{Raman spectroscopy}} 
& \multicolumn{2}{c}{\textbf{DRS}} \\
\cline{2-5}
& \textbf{Accuracy} & \textbf{F1 Score} & \textbf{Accuracy} & \textbf{F1 Score} \\
\hline
Sparse PCA + SVC & $0.681 \pm 0.023$ & $0.663 \pm 0.023$ & $0.972 \pm 0.002$ & $0.976 \pm 0.002$ \\
SVC              & $0.757 \pm 0.025$ & $0.735 \pm 0.028$ & $0.993 \pm 0.001$ & $0.994 \pm 0.001$ \\
Sparse PCA + RF  & $0.723 \pm 0.024$ & $0.705 \pm 0.026$ & $0.936 \pm 0.003$ & $0.944 \pm 0.002$ \\
RF               & $0.740 \pm 0.026$ & $0.725 \pm 0.028$ & $0.945 \pm 0.003$ & $0.952 \pm 0.003$ \\
\hline
\end{tabular}
\end{table*}

\subsection{Raman Spectroscopy Dataset}

\paragraph{Global Explanation}

\begin{figure}[!t]
    \centering
    \includegraphics[width=\textwidth]{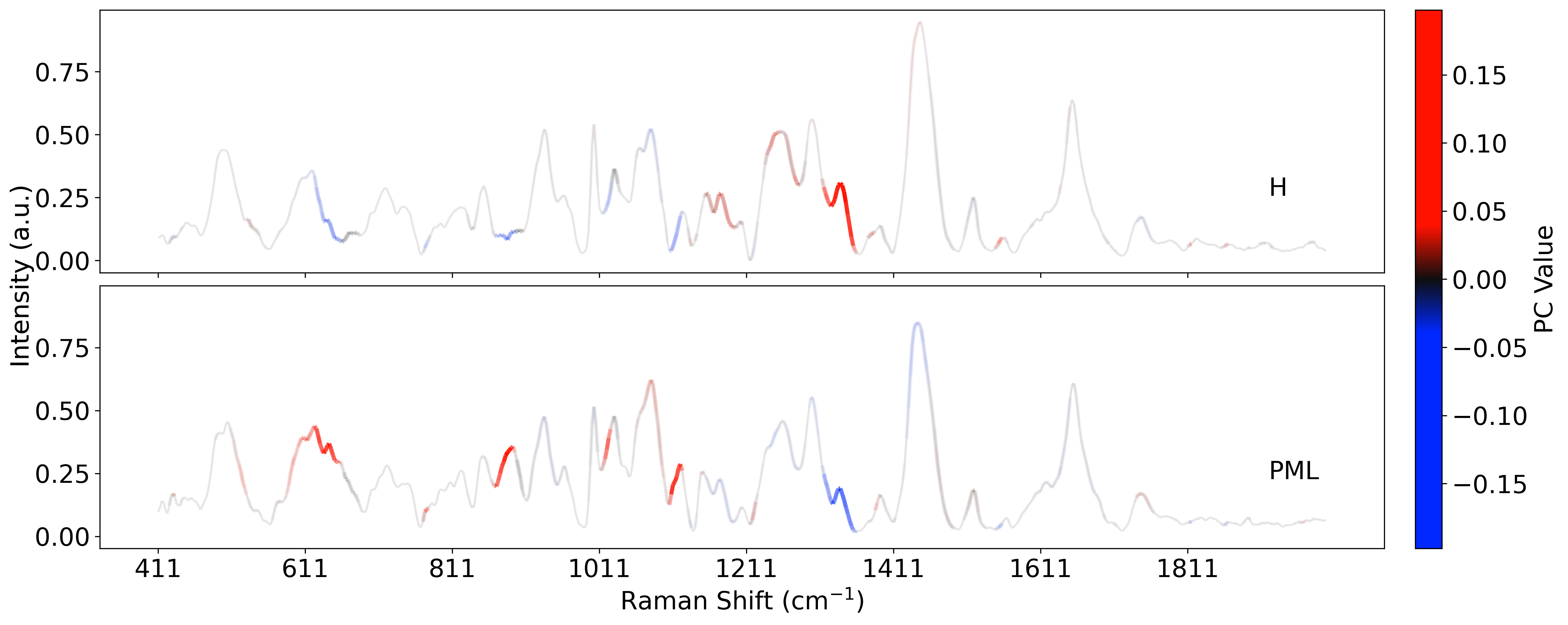}
    \caption{
     Global explanations for the binary Raman spectroscopy task (classes H and PML). The grey curves show the class-mean spectra. 
    }
    \label{fig:global_explanation_biclassification}
\end{figure}

Figure \ref{fig:global_explanation_biclassification} shows the global explanations for the binary Raman spectroscopy classification task. The grey curve corresponds to the mean Raman spectrum of each class, while the colour encodes the average normalised principal component PC values of the corresponding class projected back to the original wavenumber space. In addition, the opacity represents the averaged Shapley values, also back-projected to the original wavenumbers. 
                                                                                      
From the global explanation, several spectral regions consistently exhibit strong contributions to the model’s classification decisions, notably around 600--670~cm$^{-1}$, 860--900~cm$^{-1}$, 1010--1030~cm$^{-1}$, 1110--1130~cm$^{-1}$, 1310--1360~cm$^{-1}$. These regions therefore constitute the most decisive spectral features at the class level.

For the H class, the peaks in the 1310--1360~cm$^{-1}$ region, which are commonly associated with lipids and nucleic acids, are notably higher than the overall average spectrum, whereas the spectral intensities in the remaining regions 
600--670~cm$^{-1}$, 860--900~cm$^{-1}$, 1010--1030~cm$^{-1}$, and 1110--1130~cm$^{-1}$ are 
consistently lower than the average. In other words, these characteristic patterns serve as a key discriminative signal that the model associates with the Healthy class. In contrast, the class PML exhibits an essentially reversed trend. Consequently, when the model observes this opposite spectral pattern, it is more likely to classify the sample as PML.

% For PML samples, the spectral 
% intensity in the 1310--1360~cm$^{-1}$ region tends to be lower than the class-average spectrum, 
% while the regions around 600--670~cm$^{-1}$, 860--900~cm$^{-1}$, 1010--1030~cm$^{-1}$, and 
% 1110--1130~cm$^{-1}$ display elevated intensities.

\begin{figure}[!t]
    \centering
    \includegraphics[width=\textwidth]{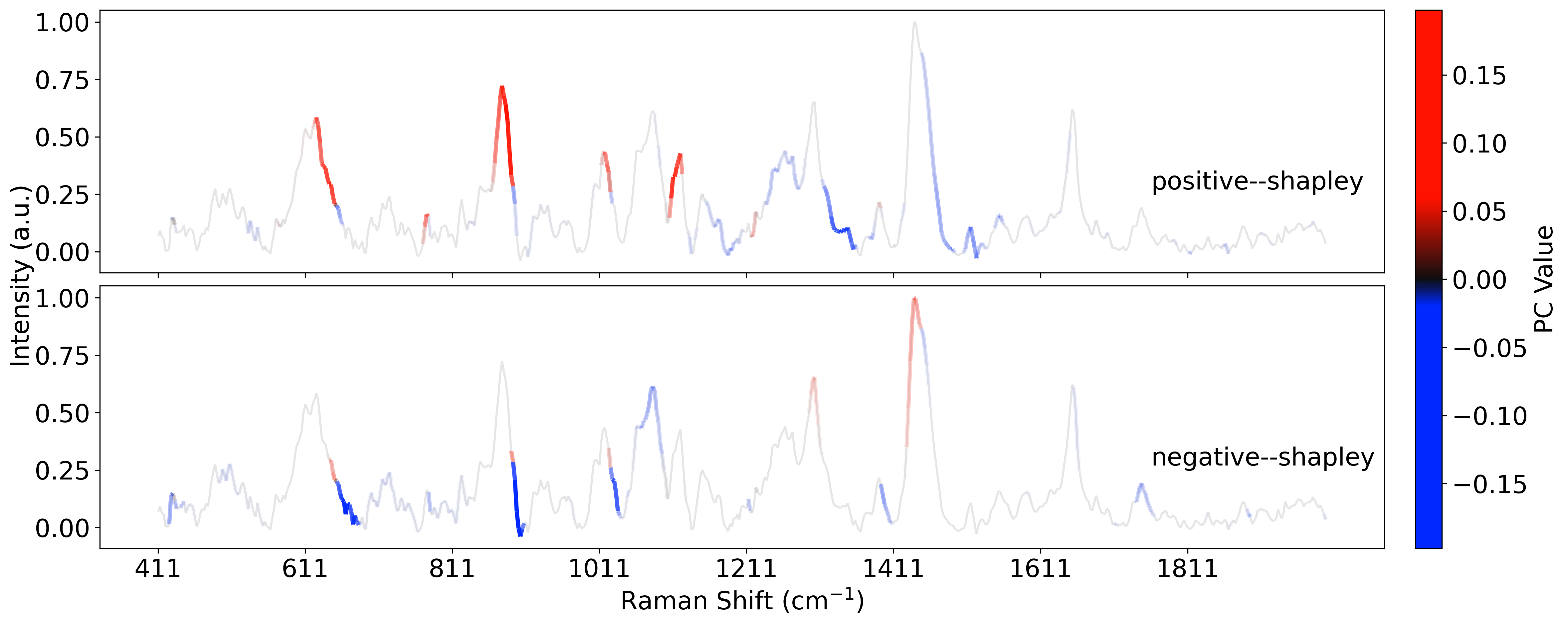}
    \caption{
    Local explanation for a binary classification instance correctly predicted as PML. The grey curve represents the Raman spectrum of the instance.
    }
    \label{fig:local_explanation_pml}
\end{figure}

\noindent\paragraph{Local Explanation}
Figure \ref{fig:local_explanation_pml} illustrates the local explanation for a single instance in the binary Raman spectroscopy classification task. The instance is labeled as PML and is correctly predicted as PML by the model. Color encodes the normalized PC value of this instance, while opacity encodes the back-projected Shapley value, which is separated into positive and negative.

In this example, elevated intensities around $620$--$640~\mathrm{cm}^{-1}$, $860$--$880~\mathrm{cm}^{-1}$, and $1110$--$1130~\mathrm{cm}^{-1}$ provide positive evidence for the PML prediction, while reduced intensity in $1310$--$1360~\mathrm{cm}^{-1}$ and $1450$--$1500~\mathrm{cm}^{-1}$ further supports this classification. In contrast, lower-than-average intensities in $640$--$670~\mathrm{cm}^{-1}$, $880$--$910~\mathrm{cm}^{-1}$ and $1020$--$1030~\mathrm{cm}^{-1}$ act against the PML label, as these patterns are more consistent with an H-like spectral profile. Overall, the model integrates both supporting and opposing evidence; here, the PML-supporting regions dominate, leading to a PML classification. Notably, the directions of these local deviations align with the class-level patterns identified in the global explanation.

\noindent\paragraph{Consistency}
Tables \ref{tab:global_consistency_corr} and \ref{tab:local_explanation_biclassification} summarise explanation reproducibility for SHAPCA and baseline SHAP using two complementary measures: cosine-based consistency and Pearson correlation. Table \ref{tab:global_consistency_corr} reports the global, class-level results and shows that SHAPCA achieves higher agreement than baseline SHAP across classes under both metrics. The SVC-based results are also markedly lower than the RF-based results, which is likely attributable to the explainer used in each case: TreeExplainer leverages the tree structure to compute SHAP values efficiently and more stably for RF, whereas KernelExplainer for SVC relies on a sampling-based approximation with limited background data as the reference distribution, making the resulting attributions more sensitive to sampling variation and considerably more expensive to compute. Table \ref{tab:local_explanation_biclassification} reports the local, instance-level results; as expected, local explanations are less reproducible than global explanations because single-sample attributions are more sensitive to noise and local decision-boundary geometry. Even so, SHAPCA continues to outperform baseline SHAP under both metrics. In particular, the local explanations produced by SVC show extremely low consistency, in some cases even yielding negative values. One possible reason is that the selected regularization setting is not well suited for explanation stability. Although the SVC hyperparameters were tuned to achieve the best predictive accuracy, the resulting model may be suboptimal from an interpretability perspective, leading to unstable local explanations across runs. Overall, these results indicate that Sparse PCA improves the stability of the resulting explanations, and that RF with TreeExplainer provides a more computationally efficient and reproducible explanation framework than SVC with KernelExplainer.

\begin{table}[H]
\centering
\caption{Comparison of global explanation consistency and correlation across classes for Raman spectroscopy data.}
\label{tab:global_consistency_corr}
\begin{tabular}{lcccccccc}
\toprule
\multirow{3}{*}{Class} 
& \multicolumn{4}{c}{SHAPCA} 
& \multicolumn{4}{c}{SHAP} \\
\cmidrule(lr){2-5} \cmidrule(lr){6-9}
& \multicolumn{2}{c}{RF+Sparse PCA} 
& \multicolumn{2}{c}{SVC+Sparse PCA} 
& \multicolumn{2}{c}{RF} 
& \multicolumn{2}{c}{SVC} \\
\cmidrule(lr){2-3} \cmidrule(lr){4-5} \cmidrule(lr){6-7} \cmidrule(lr){8-9}
& Cosine & Correlation & Cosine & Correlation 
& Cosine & Correlation & Cosine & Correlation \\
\midrule
H   & 0.820 & 0.811 & 0.458 & 0.406 & 0.715 & 0.698 & 0.370 & 0.333 \\
PML & 0.842 & 0.835 & 0.567 & 0.517 & 0.759 & 0.747 & 0.437 & 0.403 \\
\bottomrule
\end{tabular}
\end{table}

\begin{table}[H]
\centering
\caption{Comparison of local explanation consistency and correlation for Raman spectroscopy data.}
\label{tab:local_explanation_biclassification}
\begin{tabular}{lcccccccc}
\toprule
\multirow{3}{*}{} 
& \multicolumn{4}{c}{SHAPCA} 
& \multicolumn{4}{c}{SHAP} \\
\cmidrule(lr){2-5} \cmidrule(lr){6-9}
& \multicolumn{2}{c}{RF+Sparse PCA} 
& \multicolumn{2}{c}{SVC+Sparse PCA} 
& \multicolumn{2}{c}{RF} 
& \multicolumn{2}{c}{SVC} \\
\cmidrule(lr){2-3} \cmidrule(lr){4-5} \cmidrule(lr){6-7} \cmidrule(lr){8-9}
& Cosine & Correlation & Cosine & Correlation 
& Cosine & Correlation & Cosine & Correlation \\
\midrule
Local & 0.622 & 0.615 & 0.242 & 0.239 & 0.573 & 0.568 & -0.068 & -0.060 \\
\bottomrule
\end{tabular}
\end{table}

\subsection{DRS Dataset}

\begin{figure}[ht]
    \centering
    \includegraphics[width=\textwidth]{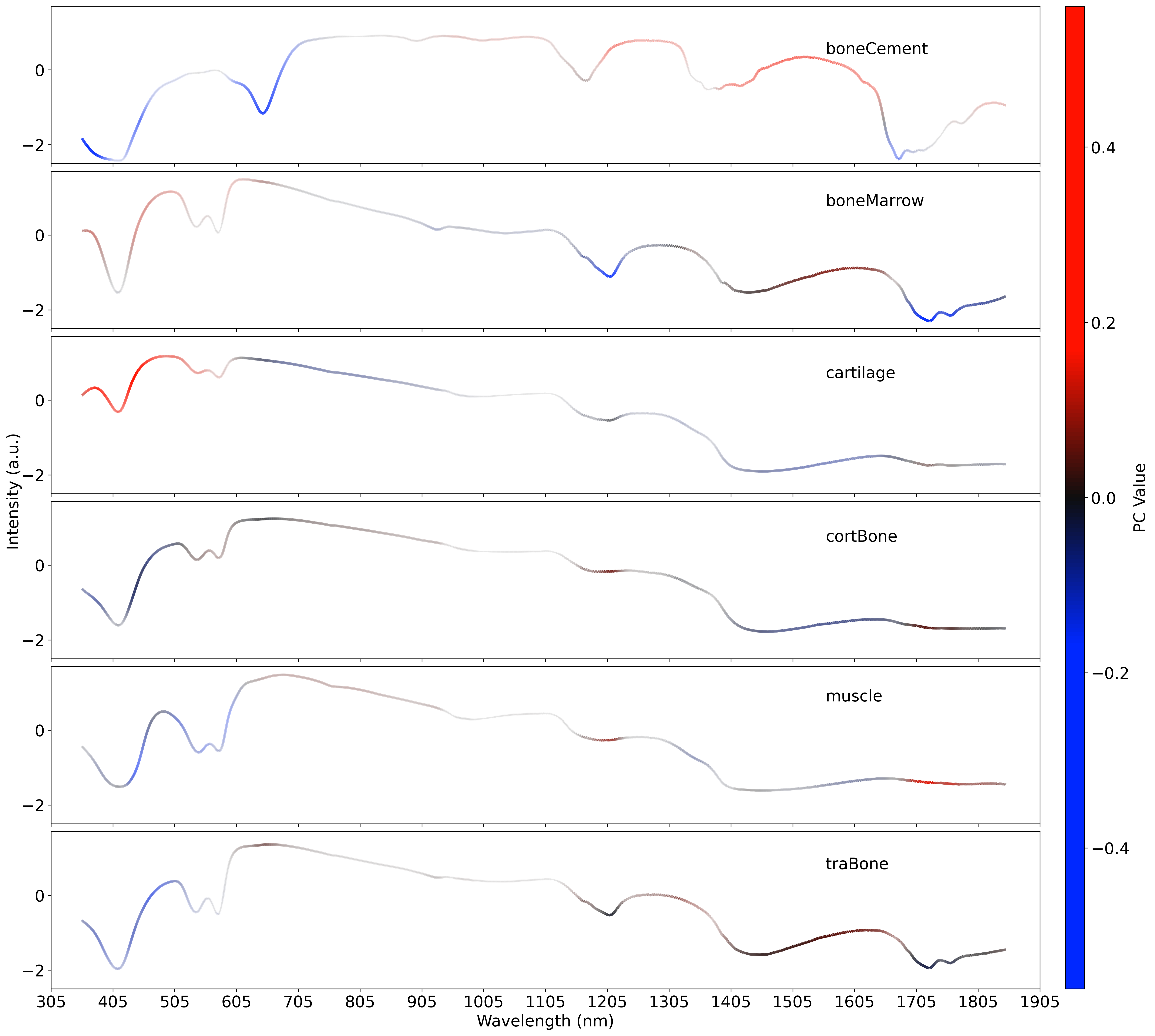}
    \caption{
    Global explanation of DRS multiclass classification across all classes. The grey curve shows the mean spectrum of each class.
    }
    \label{fig:global_explanation}
\end{figure}

\paragraph{Global Explanation}
Figure~\ref{fig:global_explanation} presents global explanations for the six classes. The following references were used as guidelines. For DRS, \cite{zhao2019minimally,dasa2018high,sekar2017diffuse} were consulted.  Notably, wavelength bands that contribute most to classification performance largely coincide with known absorption features of biological tissue chromophores, including hemoglobin, water, lipids, and collagen.  

For Bone Cement, high feature importance is observed around 400 nm and 650 nm. The band at 400 nm is likely informative because the reflectance profile in this region highlights the absence of hemoglobin, which is otherwise present in all investigated biological tissues. In contrast, the band near 650 nm is likely due to artificial pigments added to improve its visibility during surgery. This valley is clearly not present in all the biological components.

In the case of Bone Marrow, the model primarily relies on wavelength bands around 1200 nm and 1700 nm, consistent with reflectance characteristics associated with lipid absorption, reflecting the high lipid content of this tissue. Cartilage classification is dominated by the shorter wavelenghts, where its structure shows lower absorption.

For Cortical Bone, several wavelength bands contribute to classification; however, the region around 450 nm is of particular interest, as it aligns with a reflectance dip linked to the presence of oxygenated hemoglobin. Muscle tissue shows high feature importance around 420 nm relative to the shoulder of the haemoglobin peak, and the oxygenation status seems important, as the slope of the oxy and deoxy hemoglobin absorption bands in this spectral region is different, so it might differentiate more based on the shoulder. A band in 1700 nm highlighting the absence of a lipid relative to other similar tissue types 

Finally, Trabecular Bone classification predominantly relies on features between 1400 nm, 1500 nm, and 1700 nm, coinciding with a reflectance minimum primarily attributable to elevated concentrations of water and lipids.

Overall, these results indicate that while many classification-relevant features correspond to absorption characteristics of key biological components and tissue microstructure, others are selected because of their distinctive spectral contrast relative to the remaining classes.

\paragraph{Local Explanation}

\begin{figure}[!ht]
    \centering
    \includegraphics[width=\textwidth]{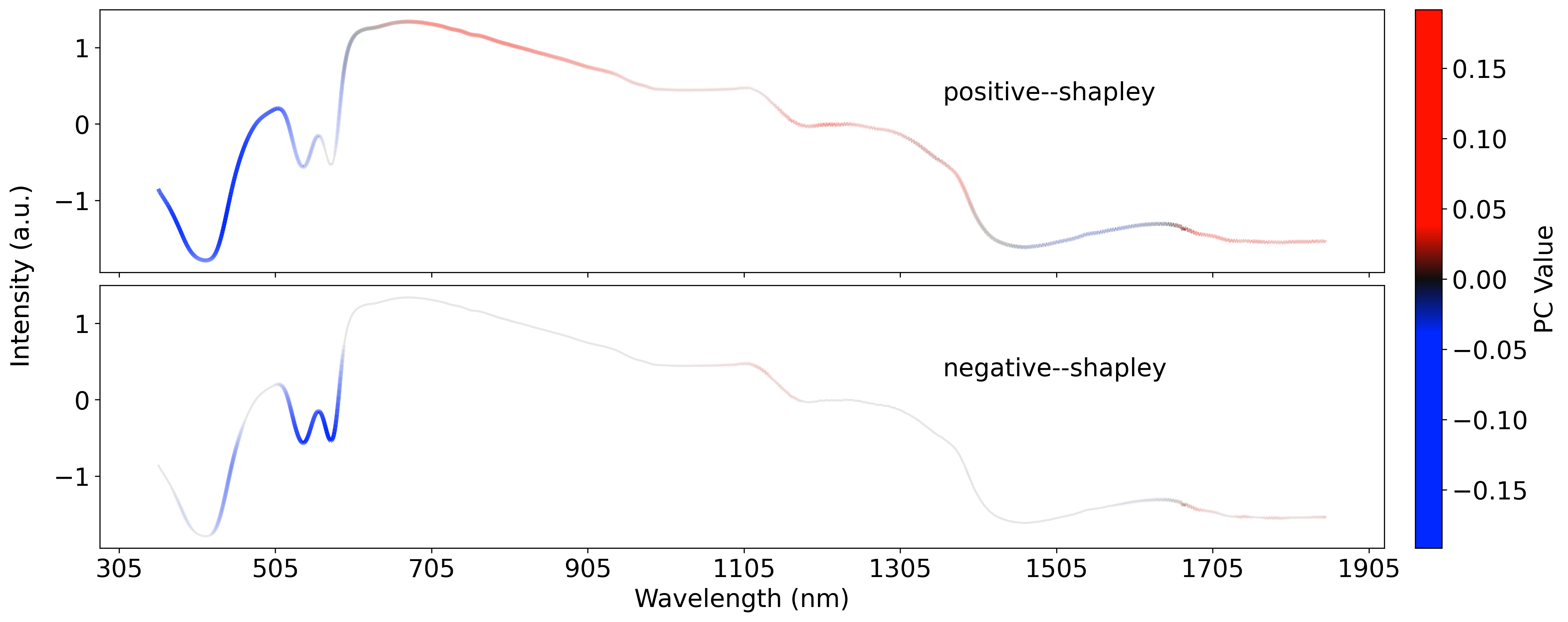}
    \caption{
    Local explanation of multi-classification for an instance correctly predicted as cortBone 
    }
    \label{fig:local_explanation_cortBone}
\end{figure}

Figure \ref{fig:local_explanation_cortBone} presents a local explanation for a single instance that is correctly classified as cortBone. The grey curve represents the DRS spectrum of the instance.

Evidence supporting the cortBone prediction includes reduced intensity in the $350$--$500~\mathrm{nm}$ region and elevated intensity in the mid-range ($705$--$905~\mathrm{nm}$). In contrast, the dominant negative evidence is concentrated around $505$--$650~\mathrm{nm}$ (blue with high opacity), indicating bands where the instance deviates toward a more muscle-like signature and thus lowers the cortBone score. Overall, the decision is governed by the strongly supportive cortBone-aligned regions; the competing short-wavelength evidence is insufficient to offset them, and the instance is therefore classified as cortBone.

\noindent\paragraph{Consistency}

Across repeated splits, SHAPCA yields substantially more reproducible explanations than baseline SHAP under both cosine similarity and Pearson correlation. For global explanations, the RF + Sparse PCA pipeline achieves consistently very high agreement across all classes, with both metrics remaining close to 1, indicating near-perfect stability. SVC + Sparse PCA is generally less stable than RF + Sparse PCA, but it still outperforms baseline SHAP in most classes. By contrast, the baseline SHAP results for RF are markedly lower and more variable, while SHAP with SVC shows the weakest overall reproducibility, despite moderate performance for a few classes such as CortBone. A similar pattern is observed for local explanations, where SHAPCA again produces substantially higher consistency than SHAP, especially for RF + Sparse PCA. Taken together, the improvements in both cosine similarity and correlation suggest that combining Sparse PCA with SHAP leads to more stable explanations across repeated splits, likely because grouping correlated spectral variables into components reduces sensitivity to sampling variation. In addition, the tree-based RF setting appears more robust than SVC in both the SHAPCA and baseline SHAP frameworks.

The consistency checks for the DRS dataset yield stronger results than those for the Raman dataset. DRS spectra have wider peaks, leading to higher collinearity between the features. In addition, the DRS dataset is considerably larger, which stabilizes model training and reduces sampling noise. As a result, when Sparse PCA estimates component directions, the extracted components are more reproducible across different random seeds and resampling runs. This improved stability propagates through the pipeline, leading to more consistent SHAP attributions and more reliable back-projected explanations.
This is particularly clear in the BoneCement, where the sample does not have the molecular variability of live tissue. In this case, the sample is considerably more reflective, providing a signature that is very different than the other classes. Consequently, many parts of the spectra can be used to discriminate it, making the SHAP explanations of the individual wavelengths quite aleatory, leading to poor consistency.

\begin{table}[H]
\centering
\small
\setlength{\tabcolsep}{3pt}
\caption{Comparison of global explanation consistency and correlation across classes for DRS data.}
\label{tab:consistency_for_global_explanation}
\begin{tabular}{lcccccccc}
\toprule
\multirow{3}{*}{Class} 
& \multicolumn{4}{c}{SHAPCA} 
& \multicolumn{4}{c}{SHAP} \\
\cmidrule(lr){2-5} \cmidrule(lr){6-9}
& \multicolumn{2}{c}{RF+Sparse PCA} 
& \multicolumn{2}{c}{SVC+Sparse PCA} 
& \multicolumn{2}{c}{RF} 
& \multicolumn{2}{c}{SVC} \\
\cmidrule(lr){2-3} \cmidrule(lr){4-5} \cmidrule(lr){6-7} \cmidrule(lr){8-9}
& Cosine & Corr. & Cosine & Corr. 
& Cosine & Corr. & Cosine & Corr. \\
\midrule
Muscle     & 0.992 & 0.984 & 0.784 & 0.787 & 0.694 & 0.679 & 0.340 & 0.280 \\
CortBone   & 0.988 & 0.948 & 0.792 & 0.706 & 0.577 & 0.576 & 0.751 & 0.748 \\
TraBone    & 0.990 & 0.976 & 0.842 & 0.735 & 0.729 & 0.694 & 0.620 & 0.607 \\
Cartilage  & 0.981 & 0.963 & 0.942 & 0.950 & 0.648 & 0.647 & 0.387 & 0.370 \\
BoneMarrow & 0.993 & 0.988 & 0.770 & 0.647 & 0.640 & 0.636 & 0.422 & 0.409 \\
BoneCement & 0.971 & 0.955 & 0.978 & 0.957 & 0.334 & 0.300 & 0.150 & 0.045 \\
\bottomrule
\end{tabular}
\end{table}

\begin{table}[ht]
\centering
\small
\setlength{\tabcolsep}{4pt}
\caption{Comparison of local explanation consistency and correlation for DRS data.}
\label{tab:local_explanation}
\begin{tabular}{lcccccccc}
\toprule
\multirow{3}{*}{} 
& \multicolumn{4}{c}{SHAPCA} 
& \multicolumn{4}{c}{SHAP} \\
\cmidrule(lr){2-5} \cmidrule(lr){6-9}
& \multicolumn{2}{c}{RF+Sparse PCA} 
& \multicolumn{2}{c}{SVC+Sparse PCA} 
& \multicolumn{2}{c}{RF} 
& \multicolumn{2}{c}{SVC} \\
\cmidrule(lr){2-3} \cmidrule(lr){4-5} \cmidrule(lr){6-7} \cmidrule(lr){8-9}
& Cosine & Corr. & Cosine & Corr. 
& Cosine & Corr. & Cosine & Corr. \\
\midrule
Local & 0.982 & 0.977 & 0.459 & 0.631 & 0.691 & 0.685 & 0.371 & 0.367 \\
\bottomrule
\end{tabular}
\end{table}

\section{Conclusion}
\label{sec:conclusion}
This work addressed the challenge of producing explanations that are stable and understandable by practitioners for machine learning models applied to spectroscopic data, where high dimensionality and strong feature collinearity often undermine the reliability of standard post-hoc explanation methods. To this end, we proposed SHAPCA, an interpretable pipeline that combines PCA with SHAP to generate feature-level explanations grounded in the underlying spectral structure.

By grouping correlated wavenumbers into sparse components, SHAPCA reduces redundancy and stabilises both model training and explanation. Performing SHAP attribution in the latent space and subsequently back-projecting contributions to the original spectral axis preserves interpretability while improving consistency across repeated runs. Experiments on Raman and DRS datasets showed that the proposed approach yields explanations that align with known biochemical and tissue-specific spectral features, while consistently outperforming baseline SHAP in terms of explanation stability, particularly at the global level.

The proposed visualisation, which jointly encodes attribution strength and relative spectral intensity, enables intuitive interpretation in terms of contiguous spectral bands rather than isolated variables, facilitating practical use by domain experts. Overall, SHAPCA provides a simple strategy for improving the robustness and interpretability of explanations in spectroscopic machine learning, supporting more trustworthy deployment in biomedical and other safety-critical applications. Future work will explore the extension of SHAPCA to other dimensionality-reduction techniques, as well as its application to longitudinal and multimodal spectroscopic datasets and other 1-D classification tasks.
% This research aims to identify stable and physically meaningful spectral regions that can reliably distinguish classes in spectroscopy data. We use Sparse PCA to group wavenumbers that vary together into a small set of sparse components, which can be linked more naturally to biochemical structure (e.g., specific chemical groups, metabolites, nucleic acids, or proteins) rather than remaining a purely mathematical construct. In this way, the learned components act as a bridge from measurements to biology, so downstream classification and SHAP-based explanations in the component space are biologically grounded, computationally efficient, and typically more consistent across runs. We then back-project both component values and SHAP attributions to the original spectral axis, producing a feature-level view where colour and opacity highlight influential regions. This representation is practical for practitioners because it emphasizes biochemical bands rather than isolated single variables, making the model’s evidence easier to relate to plausible chemical signatures. In addition, global explanations summarize the expected spectral patterns for each class, while local explanations reveal the evidence used for individual spectra; comparing the two can help flag atypical or abnormal instances.

\begin{credits}
\subsubsection{\ackname} This work was conducted with the financial support of Research Ireland - Taighde Éireann, under Grant Nos. 18/CRT/6223 and 12/RC/2289-P2, the latter being co-funded under the European Regional Development Fund. For the purpose of Open Access, the author has applied a CC BY public copyright licence to any Author Accepted Manuscript version arising from this submission.

\subsubsection{\discintname}
The authors have no competing interests to declare that are
relevant to the content of this article.
\end{credits}

% \clearpage

% \printbibliography

% \end{document}

\bibliographystyle{splncs04}
\bibliography{reference}

\end{document}